\documentclass[journal,onecolumn]{IEEEtran}
\usepackage{ragged2e}
\usepackage[a4paper, total={6.8in, 10in}]{geometry}
\usepackage{graphicx} 
\usepackage{caption}
\usepackage{setspace}
\usepackage{float}
\usepackage{xcolor}

\begin{document}

\title{Gun Detection Using Combined Human Pose and Weapon Appearance}
\pagenumbering{gobble}

\author{
\small
\begin{tabular}{c c c} 

Amulya Reddy Maligireddy&Yaswanth Reddy Parla&Manohar Reddy Uppula\\ 
Graduate Student&Graduate Student&Graduate Student\\
Rochester Institute of technology&Rochester Institute of technology&Rochester Institute of technology\\ 
am8735@g.rit.edu &yp2559@g.rit.edu&mu4587@g.rit.edu
\end{tabular}
\vspace{5pt}
\begin{tabular}{c}

Nidhi Rastogi\\ 
Assistant Professor of Software Engineering\\
Rochester Institute of technology\\ 
nxrvse@g.rit.edu
\end{tabular}
}
\maketitle

\begin{spacing}{1.1}

\section{\textbf{Introduction}}

\begin{justify}
The escalation of firearm-related incidents in recent decades has required advancements in security and surveillance systems, particularly in the identification and reduction of potential threats in public spaces. Traditional methods of gun detection, which rely mainly on manual inspections and continuous human monitoring of CCTV footage, are significantly challenged by labor intensity, high rates of false positives with high error susceptibility, where everyday objects are mistakenly identified as firearms. These challenges are exacerbated in dynamic environments such as urban centers and transport hubs, where conditions are unpredictable and constantly changing. 

\vspace{10pt}
Addressing these complexities, our research introduces a novel approach by integrating human pose estimation with weapon appearance recognition, employing advanced machine learning techniques and deep learning frameworks. This methodology aims not only to detect the presence of firearms, but also to analyze the context of their presence by examining the posture and actions of individuals potentially carrying these weapons. Traditional studies in this field have often focused on body pose estimation or weapon identification independently. For example, the methods developed by Xiao et al. [2] and Halima and Hosam [3] have laid the foundations for weapon identification using a combination of image processing and machine learning techniques. However, they typically struggle with high false positive and negative rates under the unpredictable conditions of public spaces.

\vspace{10pt}
Our initiative is to explore the possibilities of improving the accuracy and precision of the current state-of-the-art models. As part of our initial data collection, we have explored multiple open-source datasets such as IMFDB - a movie database, Monash Guns dataset, and other openly available sources that are traditionally used for real-time gun detection research. Furthermore, to improve the robustness of the model and make the data more diverse, images were created with the help of generative AI and images were manually collected from the Web with backgrounds including CCTV surveillance. Our research aims to build a robust model by training it on these real-time images to ensure generalization capacity of the model, which guarantees realistic performance evaluation.

\vspace{10pt}
The rest of the article is organized as follows, Section II discusses relevant research in the gun detection area. Section III describes the datasets utilized in this investigation. Section IV explains the Approach and Section V and VI describes the proposed architecture and methodology. Section VII summarizes the  experiments and their outcomes. Finally, Section VIII presents the conclusion and Section IX is Ethical Statement.

\end{justify}

\section{\textbf{Related Work}}
\begin{justify}
With growing cases of Gun violence, there is a need for gun detection and alerting systems. In various public places like trains, bus stations and airports, the luggage is scanned for any weapons using X-ray scanners. The images obtained from the X-ray scanners are then inspected manually for weapons which in turn carried on for long time might lead to lack of focus and lethargy \cite{1}.  A method used for automatic handgun detection from the scanned images is proposed by Xiao et al. \cite{2}, which is based on Haar-like features and AdaBoost classifiers. To make the detection economical as the X-ray scanners are expensive, gun detection in RGB images in CCTV (closed circuit television) surveillance cameras were the go-to option. There are some papers that have discussed this aspect, such as Halima and Hosam \cite{3} proposed a method to detect the presence of handguns in images. This method involved extracting SIFT (Scale-Invariant Feature Transform) features from a set of images, clustering them using the k-means algorithm, then Support Vector Machine was employed to determine whether the given image contained a weapon. Also, Tiwari and Verma \cite{4} proposed framework uses k-mean clustering algorithm to remove unrelated objects from an image by utilizing color-based segmentation. To find the gun in the segmented images, two methods are used: the Harris interest point detector and Fast Retina Keypoint (FREAK).

\vspace{10pt}
Now a days, deep learning approaches have become much more popular than standard machine learning approaches for solving computer vision problems. This is due to outstanding outcomes and their ability to automatically select the features. For an object detection problem, finding the object and its exact location would be the crucial task. There majorly two approaches for object detection, namely sliding window and region proposals approach. Sliding window involves a fixed-size window (or a bounding box) moving methodically across an image to analyze if the image has any object of interest. To predict the object in the image, there are a couple of models from existent papers. One is the Histograms of Oriented Gradients (HOG) feature descriptor \cite{5} for object detection and known for effectiveness in detecting object edges and texture information. Another is the extension of the HOG model, the Deformable Parts Models (DPM) \cite{6}. The DPM model uses HOG to detect low level features and a matching algorithm to score the sliding window by degree of match between the deformable parts’ location and the feature extracted in the image. This score is used to detect and localize the object in the image. But sliding window approach is very slow and computationally expensive. To overcome the limitations of sliding window approach, regional proposals approach is employed. In Regional proposals approach instead of using all window regions in the input image, detection proposal methods can be utilized to get the possible candidate regions. R-CNN (Region- based Convolutional Neural Networks)\cite{7} is the first of the CNNs to be utilized in this approach. It applies selective search method \cite{8} to get candidate regions which are then warped into images of same size. These images are fed to CNN- based classifier to extract features and scores the boxes using SVM. Furthermore, Fast R-CNN \cite{9}and Faster R- CNN \cite{10}have stages like ROI pooling and advantages like end-to-end training, faster and accurate object detection and better region proposals generation. Olmos et al. \cite{17}compared sliding window approaches with region-proposed methods for handgun detection on a custom-made dataset from YouTube images, finding that Faster-RCNN with a VGG16 backbone produced better results. R-CNNs, fast R-CNNs and faster R- CNNs come under two- stage detectors. Two stage detectors are accurate but not real-time compatible. 

\vspace{10pt}
Another family of detectors are single stage detectors which include YOLO (You Only Look Once)\cite{11} , SSD (Single Shot multibox detector)\cite{12}, EfficientDet\cite{13} and RetinaNet\cite{14}. There are several implementations of gun detection \cite{15}\cite{16}using YOLOv3, fine-tuned on custom datasets which have produced significant results. 

\vspace{10pt}
However, all of the discussed approaches rely on the presence of firearms in an image for gun detection without taking any external context into account. This exposes the threat for inclusion of more FPs and FNs in the predictions. In order to mitigate this, Abruzzo et al. \cite{18} have proposed an approach that uses human body posture information to estimate threat level during gun detection.  Basit et al. \cite{19} have introduced a framework that produces human-firearm paired boxes, which are then classified whether the gun is carried by human or not. However, these methods are found to be ineffective in reducing FNs as pose estimation information is added as the post processing step here, which mitigates FPs. 

\vspace{10pt}
Pose data was integrated into a handgun detector in Velasco's work \cite{22} to produce a visual rendering using heat maps that combines the handgun and the pose representation. Salido et al. \cite{21} used Open Pose \cite{20} to extract pose information to overlay it on the training images. However, this approach relies on the visual features of the pose skeletons in an image. A technique that uses body pose information to aid in the recognition of handguns was proposed by Ruiz-Santaquiteria et al. \cite{23}. In this method, the body pose keypoints and the hand region bounding boxes are extracted and fed to independent CNNs, whose decisions are then combined to get final decision. This work is later extended \cite{24} by adding data augmentation techniques based on 3D monocular pose detector and using transformer-based architectures for visual feature extraction. Distilled data-efficient Image Transformer (DeiT) \cite{25} which is the modified version of Vit \cite{26} is used in this approach. In our work, we aim to optimize the inference speed and accuracy of the above approaches by exploring latest state- of- the- art techniques.

\end{justify}
\section{\textbf{Data}}

\begin{justify}
The dataset for this study comprises of 9500 images which was constructed from multiple sources to ensure a comprehensive and representative sample. Initially, 250 images extracted from publicly available gun datasets like the Internet Movie Firearms Database (IMFDB) and the Monash Guns Dataset, see Fig.~\ref{fig:F1}, which had been previously utilized in real-time gun detection research. Also, 100 real-time CCTV images were explored from the web, which exhibited lower image quality to simulate challenging environmental conditions and enhance the robustness of the computer vision algorithms. Refer to Fig.~\ref{fig:F2} (A) for samples of the real- time CCTV images.

\begin{figure}[!htb]
    \centering
    \includegraphics[width=16.5cm]{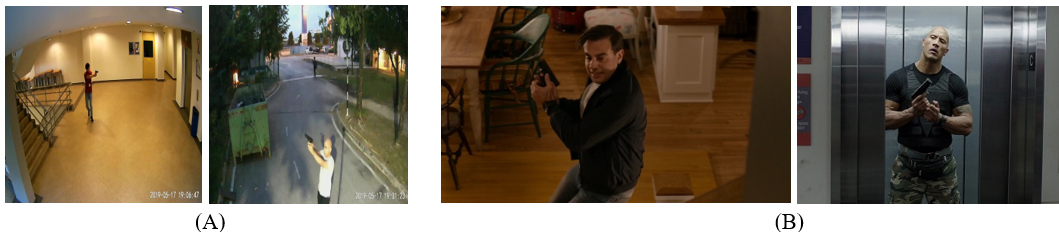}
    \caption{Sample images from (A) Monash Guns and (B) IMFDB datasets}
    \label{fig:F1}
\end{figure}

\begin{figure}[h]
    \centering
    \includegraphics[width=16.5cm]{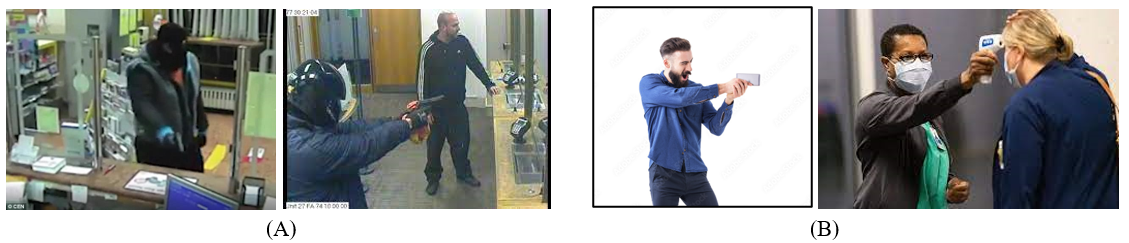}
    \caption{Samples of  (A) Real- time CCTV images (B) Benign images}
    \label{fig:F2}
\end{figure}

\begin{figure}[H]
    \centering
    \includegraphics[width=16.5cm]{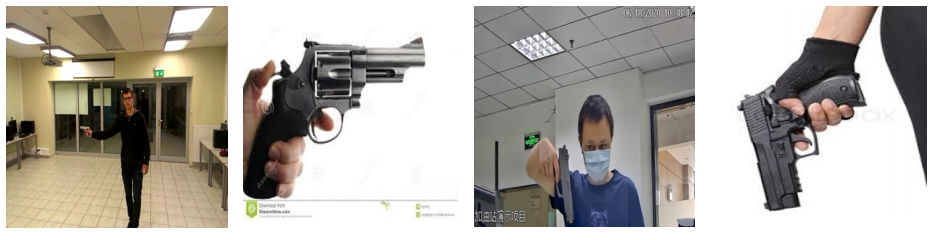}
    \caption{Samples of Images from Roboflow platfrom}
    \label{fig:F3}
\end{figure}

\vspace{10pt}
Generative AI models- Gemini Advanced were also employed to generate images depicting humans holding guns, with the aim of expanding the dataset. However, due to ethical concerns, this approach was abandoned, and through the utilization of prompts and exploratory endeavors, 100 additional images were successfully acquired from prompt outputs and suggested source links. This initiative significantly enhanced the diversity of perspectives and scenarios within our dataset.

\vspace{10pt}
The dataset also included 50 benign images, which are crucial for establishing baseline performance, understanding potential biases, and ensuring the robustness of the machine learning and computer vision models in real-world applications.
Refer to Table \ref{table:1} for the summary count of images aggregated from various datasets in our final dataset.

\begin{table}[h!]
\centering
\begin{tabular}{ |c|c| }
\hline
\textbf{Datasource} & \textbf{Number of Images}  \\
\hline
{IMFDB} & 150  \\ 
{Monash Guns}& 100 \\ 
{Real-time CCTV images}& 100\\ 
{GenAI prompts}& 100\\ 
{Benign}& 50\\
{Roboflow}& 9000\\
\hline\hline
\textbf{Total Images}& 9500\\
\hline
\end{tabular}
\caption{Image counts from diverse datasets in our final dataset}
\label{table:1}
\end{table}

\vspace{10pt}
To ensure accurate model training and evaluation, these images were meticulously annotated using the VGG Image Annotator (VIA) tool. This manual annotation process enabled precise labeling of relevant objects within the dataset, providing a robust foundation for supervised learning tasks. Subsequently, for the integration of pose estimation alongside object detection, the images were re-annotated in the YOLO format, facilitating compatibility with the requirements of advanced detection and estimation models. 

\vspace{10pt}
In the next stages of project, additionally the final dataset for this study was sourced from Roboflow, a platform designed to facilitate the creation, annotation, and preprocessing of computer vision datasets. The dataset comprises 9,000 images, with guns annotated in the YOLOv8 format, a widely adopted standard for efficient object detection tasks. Preprocessing involved two key steps. First, auto-orientation was applied to correct pixel alignment issues based on EXIF metadata, accompanied by the removal of EXIF-orientation tags to ensure consistent image presentation. Second, all images were resized to dimensions of 640x640 pixels using a stretch method, which standardizes the input size while modifying the original aspect ratio. These preprocessing steps were undertaken to ensure uniformity and compatibility for training object detection models, enhancing the dataset's utility in downstream tasks. Refer to Fig.~\ref{fig:F3} for data samples from Roboflow platform.

\vspace{10pt}
{\textbf{Image Augmentation:}}
While larger datasets can enhance model performance, augmenting existing images significantly improves the diversity and robustness of the training data. Image augmentation techniques such as rotation, scaling, flipping, and color adjustments facilitate the creation of varied perspectives and conditions from a limited set of images. This approach not only mitigates the risk of overfitting but also enables the model to generalize more effectively across different scenarios. To improve overall generalization across varied conditions, augmentation techniques—including horizontal flipping, rotation, scaling, hue adjustment, saturation adjustment, and brightness adjustment—were integrated into the data.yaml file. The image augmentation parameters are mentioned in Fig.~\ref{fig:F4}

\begin{figure}[h]
    \centering
    \includegraphics[width=16.5cm]{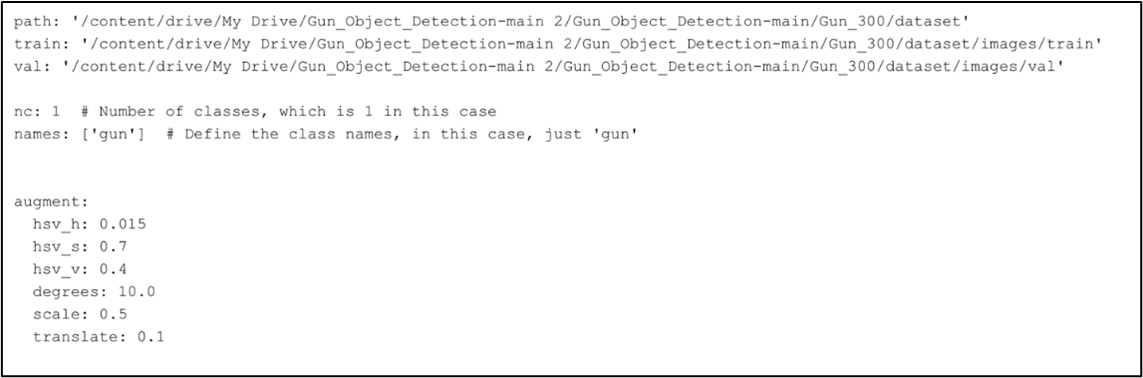}
    \caption{Image Augmentation code snippet with parameters}
    \label{fig:F4}
\end{figure}

\end{justify}
\section{\textbf{Approach}}
\begin{justify}
The major issues in the detection of weapons from images or videos are occlusion and the background objects have gun-like structure. Occlusion can occur due to many reasons: (1) The gun is held with both hands and one of the hands covers the gun, (2) Gun is shadowed by other humans or objects (3) Poor image quality, and this would lead to false- negatives (FNs). FNs have very severe consequences as the weapons are undetected in actual violence situation. The gun look- alike objects in the background will result in false- positives (FPs), which in turn lead to unnecessary public panic and mistrust in the system.  
\captionsetup{justification=centering}
\begin{figure}[h]
    \centering
    \includegraphics[width=11cm]{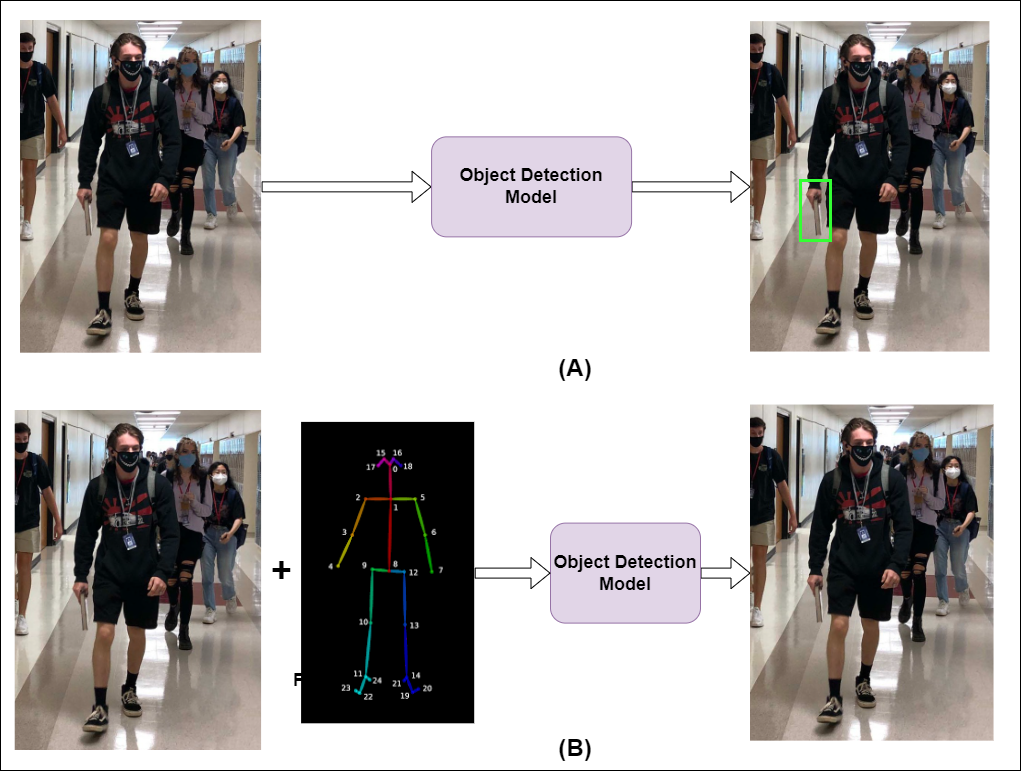 }
    \caption{False Positive correction (A) without human pose (B) with human pose information}
    \label{fig:F5}
\end{figure}

\begin{figure}[h]
    \centering
    \includegraphics[width=11cm]{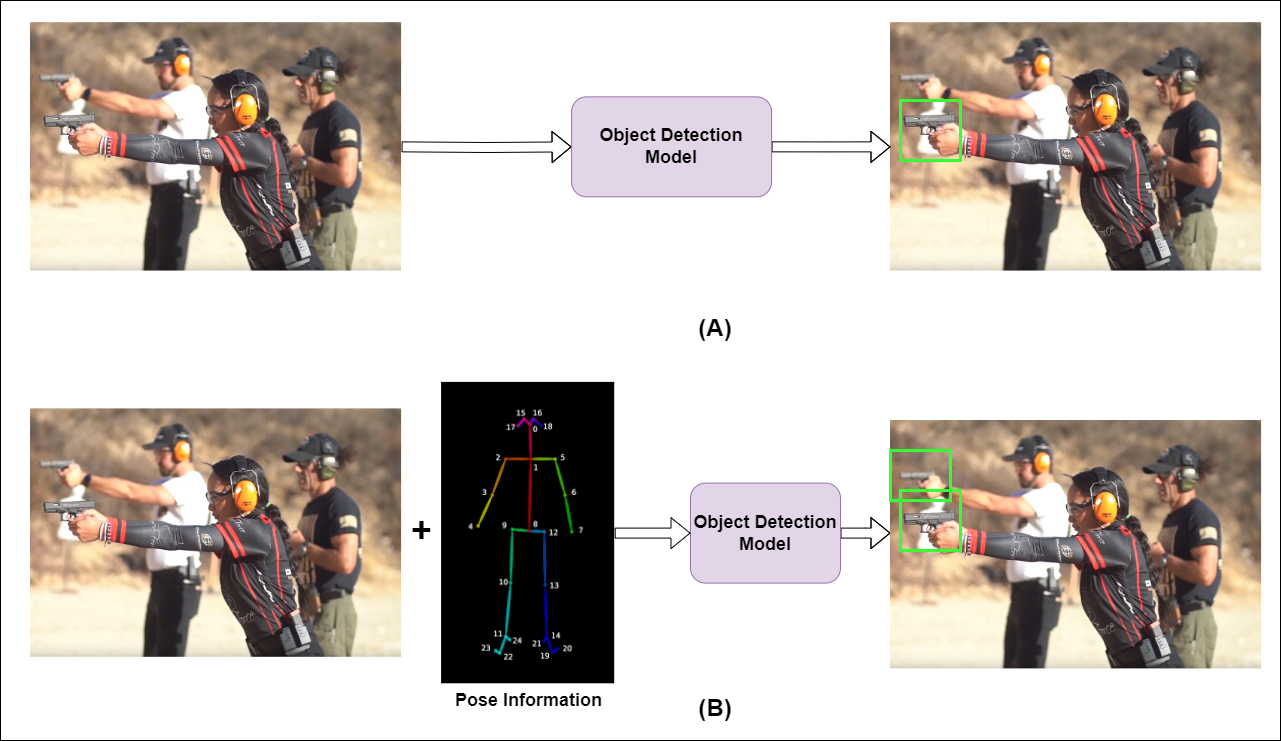 }
    \caption{False Negative correction (A) without human pose (B) with human pose information}
    \label{fig:F6}
\end{figure}

\vspace{10pt}
To mitigate these FNs and FPs, integrating human pose with weapon appearance will give more contextual understanding and accuracy in gun identification related to threat. There is a distinctive posture when holding a gun which will stand out from other common activities. In the Fig.~\ref{fig:F5}, (A) indicates false positive where the student is holding a book, and it is recognized as a gun without human pose information and (B) shows how the book is not detected as gun with human pose information. The scenario of false negatives is shown in Fig.~\ref{fig:F6}, where in (A) the rear end person is holding a gun but not detected due to poor quality of image, in (B) gun is predicted by incorporating human pose data.

\vspace{10pt}
In the research of gun detection and incorporating human pose information, there are numerous state- of- the- art approaches to detect carried gun with most confidence like: (1) Estimating body pose key points and detecting bounding box and pass them through combined detection mask \cite{22}, (2) Imposing the estimated body pose key points onto the original input image and processing it forward to classify the carried gun scenario \cite{21} and, there is also need for threshold tuning and sensitive learning. 

\vspace{5pt}
These methods have their own limitations. But considering them as the baseline approaches, moving forward, we plan to focus on exploring cutting-edge techniques in computer vision and object detection to enhance accuracy, computational efficiency, and reliability. This involves investigating novel approaches for incorporating body pose information with visual features within an image to achieve more efficient gun detection. Specifically, we aim to combine the extracted visual features and body pose keypoints to create a comprehensive feature representation, which will then be fed into the object detection model.

\vspace{5pt}
To improve the accuracy of our firearm recognition technology, we are compiling a varied dataset that contains images of people in a variety of poses, both with and without firearms from internet. This includes challenging scenarios, such as low-quality CCTV footage in which firearms may be disguised by other objects or hidden in bad lighting conditions. This large and diversified dataset is critical in training the algorithm to efficiently identify firearms under a variety of real-time conditions, considerably improving its dependability and accuracy aiming to reduce the false positive and false negatives.
\end{justify}

\section{\textbf{Architecture}}
\begin{justify}

\begin{figure}[h]
    \centering
    \includegraphics[width=16.5cm]{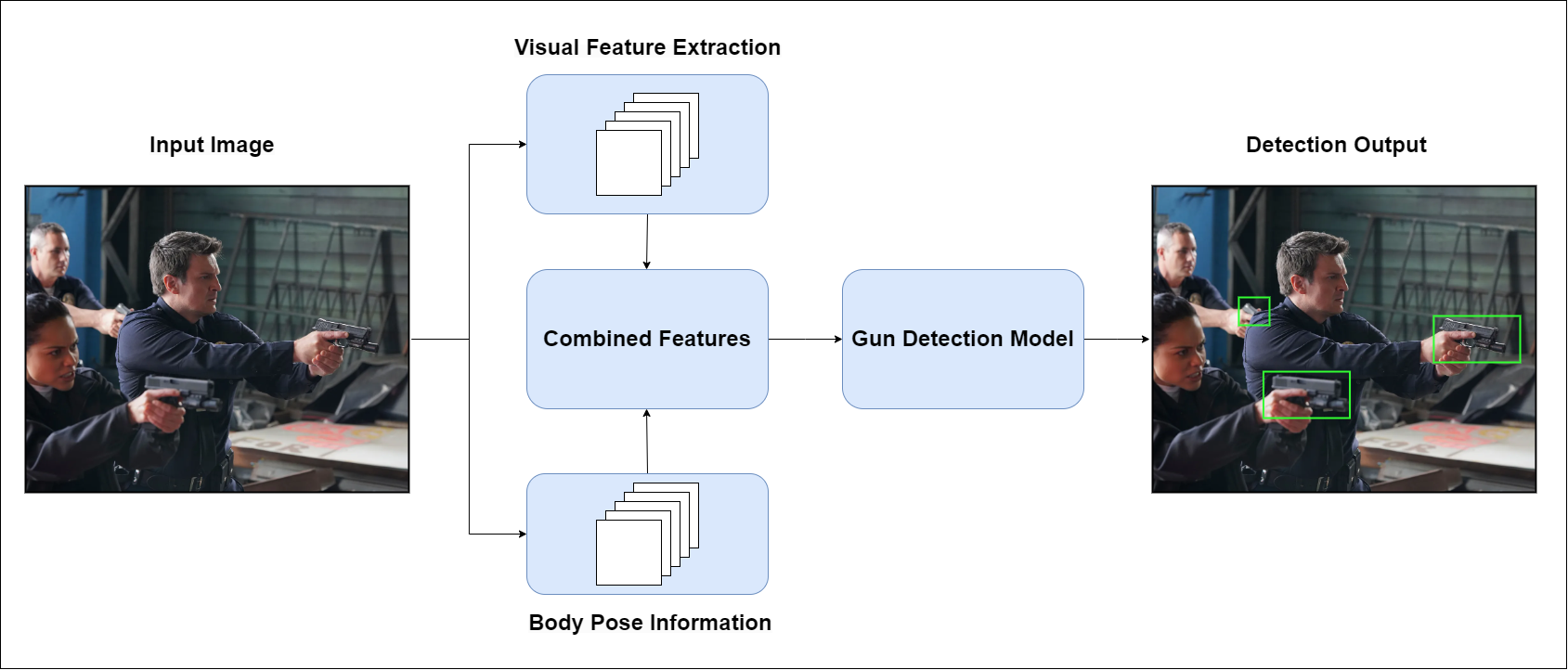}
    \caption{Proposed Architecture}
    \label{fig:F7}
\end{figure}
Our proposed gun detection model architecture is shown visually in Fig.~\ref{fig:F7} . Initially the input image is fed into CNN architecture to extract visual features and a pose estimations framework is used to obtain the pose information of each individual present in an image. These features are combined to pass through an object detection model to obtain the final detection of guns in an image. Each individual component of the architecture is explained in detail below. 

\vspace{10pt}
\textbf{Visual Feature Extraction:}
In this step, Convolutional Neural Network and Transformer architectures are employed to extract visual features from a raw image related to presence of a gun. Low level features like edges, texture or patterns are extracted by applying filters initially in the convolutional layers. More complex features are captured in the next stages of the network leading to the formation of high-level visual representation of an image.  

\vspace{10pt}
\textbf{Body Pose Estimation: } 
Body pose analysis of an individual holding a gun can offer valuable insights for effective firearm detection in images. Various pose estimation architectures can be employed for this task, allowing for the identification of key body joints and their spatial relationships with objects.
These architectures typically utilize deep neural networks to detect key body joints such as shoulders, elbows, wrists, hips, knees, ankles, and the face. Confidence maps are generated for each keypoint, indicating the likelihood of their presence at specific locations in the image. These confidence maps are then used to localize the positions of body joints within the input image, providing 2D keypoint coordinates.
These 2D keypoint coordinates can be leveraged to understand the spatial relationship between body joints and objects within the scene, aiding in firearm detection.

\vspace{10pt}
\textbf{Gun Detection:}
The extracted visual features and body pose keypoints are combined to create a comprehensive feature representation, which is then fed into object detection model. Widely used object detection models like YOLO, Faster-RCNN and Detection-Transformers will be used after fine-tuning specifically for gun detection in our approach. The final output from this model will be the image with the bounding-box prediction of presence of a gun with a confidence score. 
\end{justify}

\section{\textbf{Methodology}}

\subsection {\textbf{Baseline Object Detection Model}}
\begin{figure}[!h]
    \centering
    \includegraphics[width=11cm]{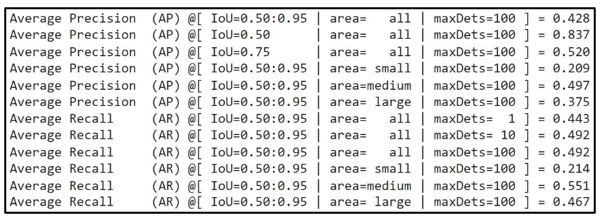}
    \caption{Baseline object detection model performance metrics}
    \label{fig:F8}
\end{figure}

The initial phase involved establishing a baseline object detection model using the Faster R-CNN architecture with a ResNet50 backbone pretrained on the COCO dataset. The model was trained on a dataset of 400 images and validated using 100 images (Roboflow images were used in the later stages of the project). This foundational model achieved a moderate mean Average Precision (mAP) of 0.410, but performance for smaller objects, with an Average Precision (AP) of 0.153, highlighted challenges such as detecting partially occluded firearms, objects held in the background, or objects of similar size and shape. These findings, see Fig.~\ref{fig:F8}, informed the need for subsequent improvements.
 
\begin{figure}[!h]
    \centering
    \includegraphics[width=16.5cm, height=5cm]{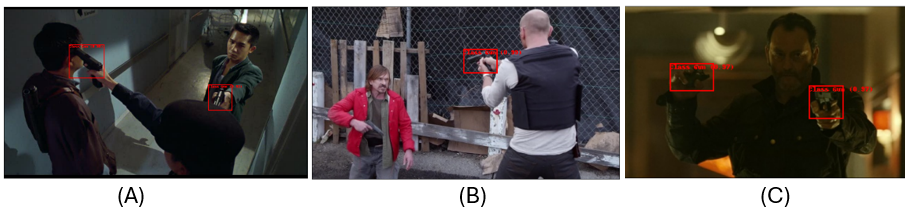 }
    \caption{Baseline object detection model performance on IMFDB images (A) True Positive (B) False Negative (C) False Positive}
    \label{fig:F9}
\end{figure}

\begin{figure}[!h]
    \centering
    \includegraphics[width=16.5cm]{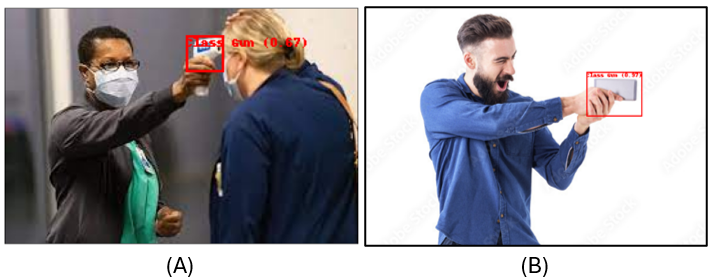}
    \caption{Benign Images' object detection in cases of a person holding (A) Infrared Thermometer (B) Mobile phone}
    \label{fig:F10}
\end{figure}

\vspace{10pt}
In Fig.~\ref{fig:F9}, the baseline object detection model successfully detects the gun. However, instances of false positives (FPs) and false negatives (FNs) are evident.  Furthermore, as shown in Fig.~\ref{fig:F10}, (A)
and (B), although the individuals hold infrared thermometer and a mobile phone, the
baseline object detection system misinterprets them as firearms, leading to incorrect categorization.

\subsection{\textbf{Pose Estimation for Enhanced Detection}}
To address the baseline model’s limitations, human pose estimation can be integrated to complement object detection by leveraging the spatial relationships of key body joints (e.g., elbows, wrists, knees) for contextualizing firearm presence. Several pose estimation models were evaluated, including OpenPose, MoveNet, AlphaPose, and YOLOv8. 

\vspace{10pt}
YOLOv8 emerged as the optimal choice due to its single-stage object detection pipeline with integrated top-down pose estimation, offering an efficient and accurate approach suitable for diverse scenarios.
The YOLOv8 family includes variants such as Nano (n), Small (s), Medium (m), Large (l), and Extra Large (x). For this project, YOLOv8n (Nano) and YOLOv8s (Small) were selected for their balance of speed, accuracy, and computational efficiency, making them suitable for real-time applications in resource-constrained environments.

\vspace{10pt}
\textbf{Comparative YOLOv8 variants Performance Analysis:}
The training of YOLOv8n and YOLOv8s was conducted with the following configurations:

\vspace{10pt}
\begin{itemize}
    \item Input Dataset: 450 augmented images
    \item Hyperparameters:
        \begin{itemize}
            \item Number of epochs: 35
            \item Learning rates: 0.01, 0.005, 0.001
            \item Optimizers: AdamW, SGD
            \item Image sizes: 1024, 640
            \item Batch sizes: 16, 8
        \end{itemize}

\end{itemize}

\begin{figure}[!h]
    \centering
    \includegraphics[width=16.5cm]{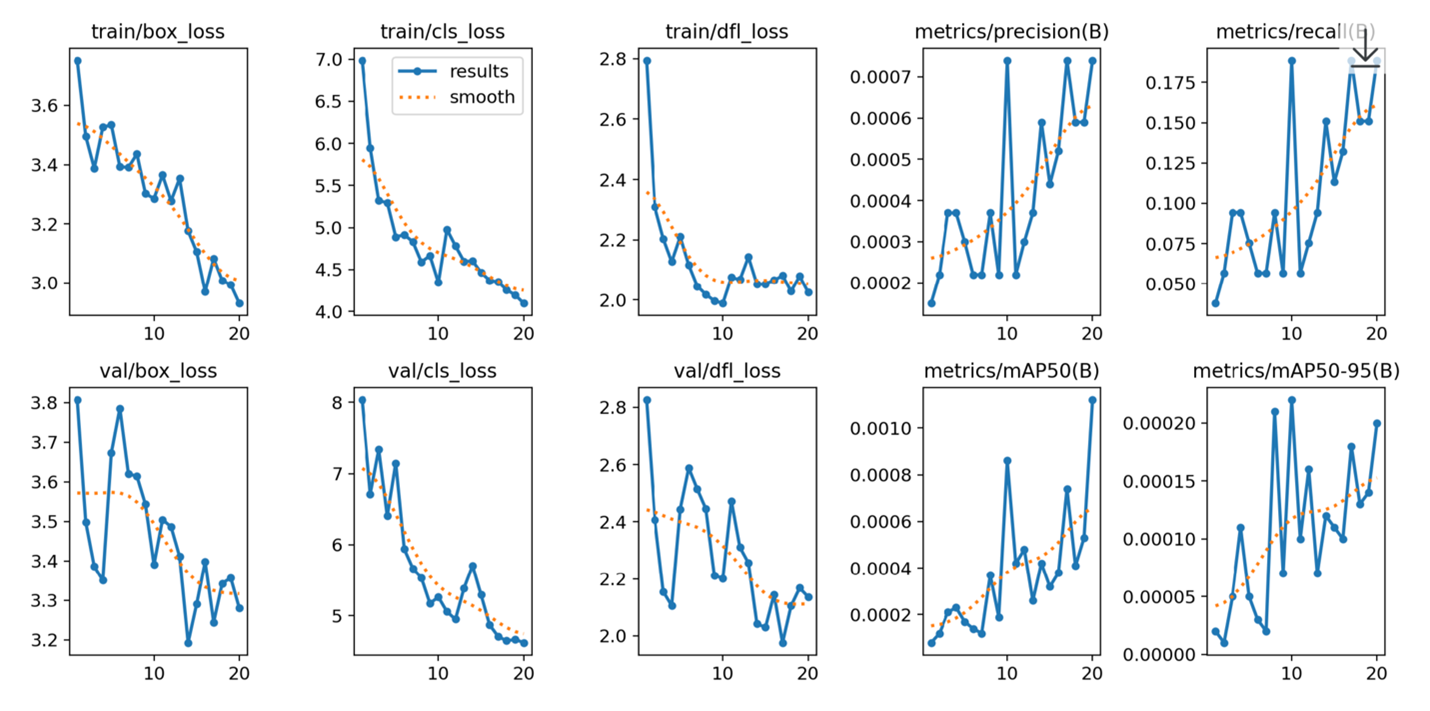}
    \caption{Plots of YOLOv8n Perfomance metrics}
    \label{fig:F11}
\end{figure}

\begin{figure}[!h]
    \centering
    \includegraphics[width=16.5cm]{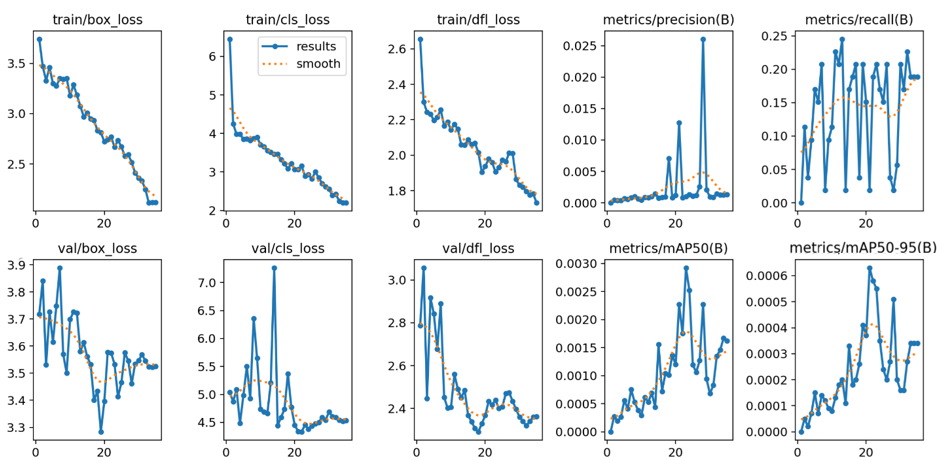 }
    \caption{Plots of YOLOv8s Perfomance metrics}
    \label{fig:F12}
\end{figure}

\vspace{10pt}
Refer to Fig.~\ref{fig:F11} and Fig.~\ref{fig:F12}  for plots of performance metrics of YOLOv8n and YOLOv8s models respectively. The performance comparison revealed that while YOLOv8n excelled in fast inference with lower resource demands, YOLOv8s consistently outperformed YOLOv8n across key metrics. YOLOv8s demonstrated: 

\vspace{10pt}
\begin{itemize}
    \item Lower training and validation box and classification losses, indicating improved object localization and classification accuracy.
    \item Higher stability in precision and recall metrics, reducing fluctuations observed in YOLOv8n.
    \item Superior mAP scores, including mAP50 and mAP50-95, reflecting enhanced detection accuracy at varying IoU thresholds.
\end{itemize}

\vspace{10pt}
These results highlight YOLOv8s as the preferred choice for applications requiring higher accuracy, despite its slightly higher resource requirements compared to YOLOv8n.

\begin{figure}[!h]
    \centering
    \includegraphics[width=16.5cm]{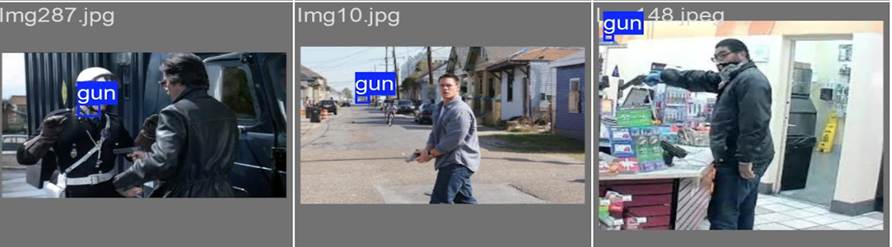}
    \caption{YOLOv8s Model's Performance on test images}
    \label{fig:F13}
\end{figure}

\vspace{10pt}
The analysis of the test images collation in Fig.~\ref{fig:F13} reveals that, although the gun is detected, the placement of the bounding box is inaccurate. This misalignment may be attributed to several factors, including variations in the orientation of the gun, potential occlusion by surrounding objects, or limitations in the training dataset that did not adequately represent similar scenarios. 

\vspace{10pt}
The observed challenges in accurately localizing objects are likely attributed to the limited diversity and representativeness of the initial training dataset. To mitigate this issue, the dataset was expanded with 9,000 additional images sourced from Roboflow. While this enhancement significantly improved the object detection capabilities of the YOLOv8s model, it did not facilitate reliable human pose estimation. To address this limitation, MediaPipe was integrated for pose estimation, particularly hand keypoint position extraction, complementing YOLOv8s, which continued to be utilized for object detection.

\vspace{10pt}
To further refine the firearm detection system, hyperparameter tuning was conducted to optimize the performance of YOLOv8s for object detection and MediaPipe for pose estimation.. A dynamic proximity threshold was introduced, adjusting the hand-to-gun distance threshold based on gun size to account for variations in bounding box dimensions. YOLOv8s confidence thresholds were also lowered to detect partially occluded or low-confidence objects effectively. Additionally, visual annotations of detected landmarks and bounding boxes were incorporated to facilitate debugging and interpretability, ultimately refining the system's ability to analyze spatial relationships between firearms and human body parts for accurate threat detection.

\vspace{10pt}
\textbf{Threat Classification Logic:}
To determine potential threats, the system applied a series of rules based on the spatial relationships between detected hands and firearms:

\begin{enumerate}
    \item Calculated Euclidean distances between gun centers and hand landmarks.
    \item Applied dynamic thresholds proportional to the gun size (e.g., 50\% of the bounding box width).
    \item Flagged a threat if the hand-to-gun distance fell below the threshold.
    \item Annotated detected threats on images with a ``Threat Detected!" label and highlighted guns with red bounding boxes.
\end{enumerate}

\section{\textbf{Results}}

\subsection{\textbf{Object and Human Detection}}
\begin{figure}[!h]
    \centering
    \includegraphics[width=16.5cm]{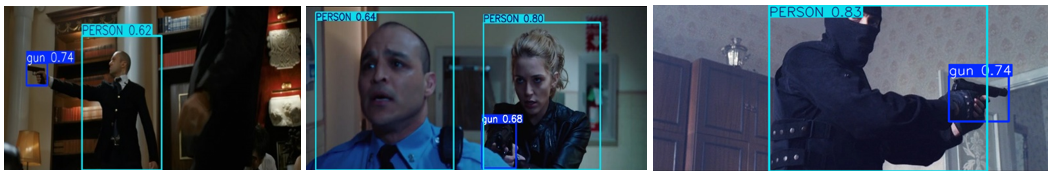}
    \caption{Images with bounding boxes for both guns and persons}
    \label{fig:F14}
\end{figure}

The initial dataset contained limited images focusing solely on gun detection, which constrained the YOLOv8s model's ability to generalize and accurately detect related objects in realistic scenarios. By augmenting the dataset with 9,000 images from Roboflow, including bounding boxes for both guns and persons (Refer Fig.~\ref{fig:F14}), the model achieved a significant improvement in detection performance. This enhancement allows the detection of guns in the context of their proximity to persons, which is critical for subsequent applications such as pose estimation and threat assessment. The inclusion of bounding boxes for persons has enabled the model to jointly detect guns and individuals, establishing a critical foundation for multi-class object detection in safety-critical environments. 

\vspace{10pt}
\subsection{\textbf{Integration of Mediapipe's Pose module with YOLOv8s' object detection}}
\begin{figure}[!h]
    \centering
    \includegraphics[width=16.5cm, height= 5cm]{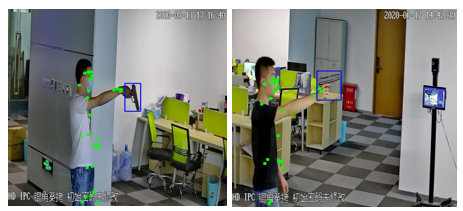}
    \caption{Images integrated with Mediapipe's body keypoints and YOLOv8s' object(gun) detection}
    \label{fig:F15}
\end{figure}

The provided images in Fig.~\ref{fig:F15}, showcase the integration of Mediapipe's Pose module with YOLO-based object detection to identify human pose keypoints alongside detecting guns. Mediapipe captures keypoints across the body, including arms, hands, and torso, while YOLO localizes guns with bounding boxes. This integration enables analyzing the spatial relationship between a person's body parts and the gun, offering contextual insights such as whether the gun is being held or aimed. Such a system is crucial for security applications, where understanding human pose in relation to weapons enhances threat detection and assessment.

Building on the final model with the integration of Mediapipe's Pose module with YOLO for gun detection, additional layers of analysis were incorporated to enhance threat assessment. Hand detection leverages Mediapipe's pose keypoints to identify hand coordinates and checks for overlaps with the detected gun's bounding box. Using the gun's bounding box and the direction of the person's pose, the system infers whether the gun is aimed or in a potentially threatening position. Threat classification is further refined by defining rules based on the distance between the gun and the hand, the arm's orientation relative to the gun, and the gun's position relative to the torso or head. Mediapipe Pose processes the cropped images of detected individuals to map key landmarks, including hands, elbows, and shoulders, enabling precise analysis of body posture. 

\begin{figure}[!h]
    \centering
    \includegraphics[width=16.5cm]{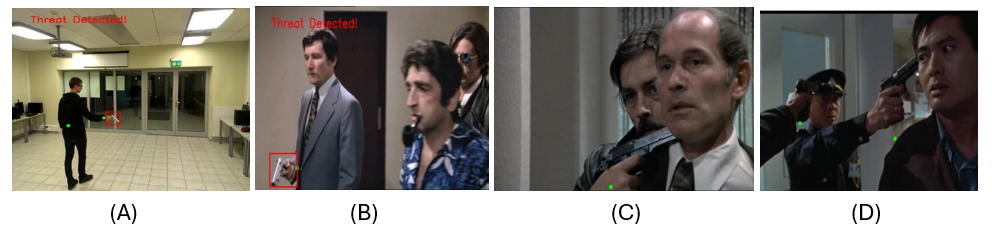}
    \caption{Output of the final model with threat classification logic}
    \label{fig:F16}
\end{figure}

\vspace{10pt}
Fig.~\ref{fig:F16} depicts the output of the final model built with threat classification logic. The system successfully identifies guns in the environment by enclosing them in bounding boxes and marking their proximity to human body parts. Key points like hands and wrists, detected via pose landmarks, are crucial to analyzing the relationship between the gun and the individual holding it. Please refer to the threat classification logic discussed in the methodology section of this report. For instance in Fig.~\ref{fig:F16} (A), (B) , the "Threat Detected!" labels signify scenarios where guns are in close proximity to critical body parts, such as hands, thereby indicating a potential threat. The inclusion of these visual cues highlights the model's ability to assess threat contexts effectively. However, in Fig.~\ref{fig:F16} (C) and (D) images show only one detected body point (e.g., the hand or wrist), which reflects the partial detection accuracy due to occlusions or pose complexities. These cases of missed detections (False Negatives), emphasizing the scope for further optimization.

\begin{figure}[!h]
    \centering
    \includegraphics[width=16.5cm]{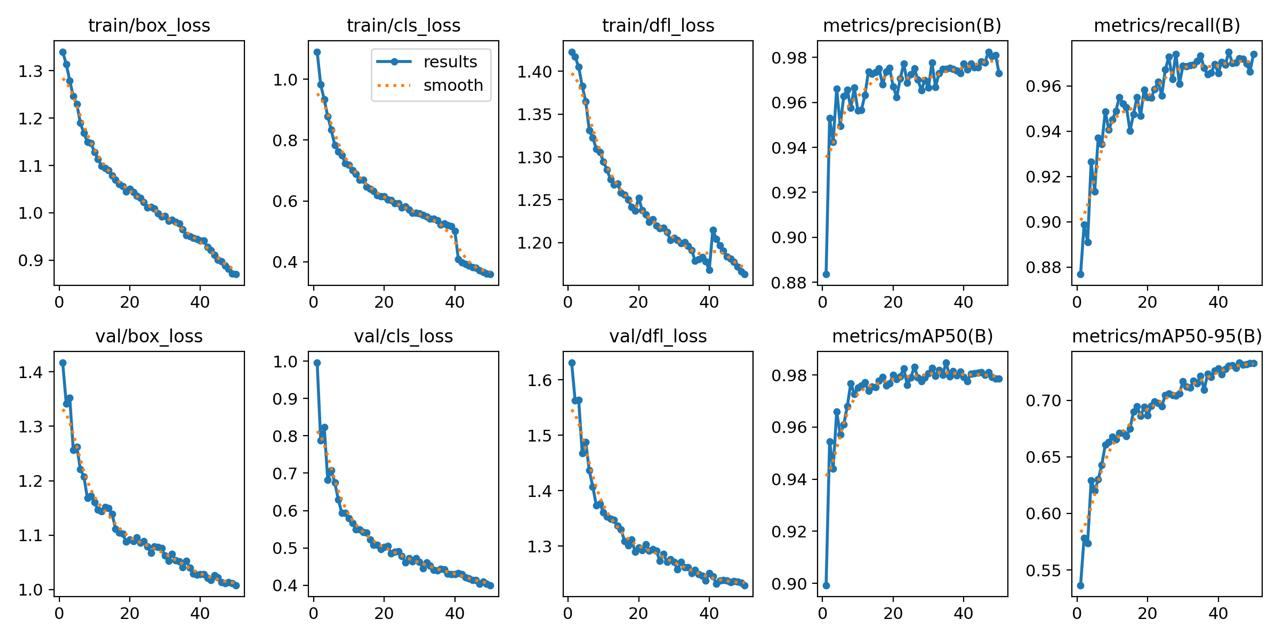}
    \caption{Performance Metrics plots of the final model}
    \label{fig:F17}
\end{figure}

\vspace{10pt}

Refering to Fig.~\ref{fig:F17}, from plots of performance metrics of the final model integrated with YOLOv8s model and Mediapipe Pose estimation shows notable improvements over the previous version of using just YOLOv8s model, achieving better optimization, generalization, and accuracy. Here are the key enhancements observed:
\begin{itemize}
    \item The final model significantly reduced training losses (box, classification, and DFL), indicating better optimization and Validation losses are now smoother and more stable, showing improved generalization
    \item Precision and recall have increased drastically, reflecting fewer false positives and negatives.
    \item The current model achieves much higher mAP scores (mAP50 ~0.98, mAP50-95 ~0.70), indicating better object detection accuracy.
    \item Faster and smoother convergence shows the updated model trains more efficiently within fewer epochs.
\end{itemize}

\vspace{10pt}
While these improvements are substantial, further enhancements can be made to optimize performance in edge cases and real-world scenarios.

\section{\textbf{Conclusion}}
\begin{justify}
The results of our study confirm that the integration of human pose information with weapon appearance creates a tool to enhancing public safety through improved gun detection. Our system's ability to discern subtle nuances in human posture associated with weapon carriage brings a novel aspect to surveillance technology, significantly reducing the likelihood of false identifications.

\vspace{10pt}
The project integrates YOLOv8s model for gun detection and Mediapipe's pose estimation for analyzing human poses, culminating in a robust system for threat detection based on proximity between guns and critical body landmarks like hands or wrists. By leveraging a dataset of 9,500 images with bounding boxes for guns and humans, the model achieves notable detection accuracy for identifying guns and contextualizing their position relative to human poses. Threats are flagged when guns are in proximity to key body parts, effectively enhancing situational awareness. The model's ability to combine object detection with human pose estimation demonstrates its potential application in public safety, security systems, and real-time surveillance scenarios. Furthermore, the integration of distance thresholds and dynamic bounding box adjustments improves adaptability across diverse environments and scales of threats. The iterative inclusion of data augmentation and hyperparameter tuning significantly contributed to improving detection accuracy and model generalization.

\subsection{\textbf{Future Scope}}
\begin{enumerate}
    \item Enhanced Threat Classification
    \begin{itemize}
        \item Incorporate advanced rules or train neural networks to classify poses as threatening or non-threatening based on nuanced arm positions, gun orientations, and contextual movement.
        \item Expand the criteria for threats by including multi-object scenarios, such as distinguishing between aggressors and victims in complex scenes.
    \end{itemize}
    \item Dataset Expansion
    \begin{itemize}
        \item Further augment the dataset to include diverse environmental conditions, camera angles, and scenarios involving occlusions or crowd settings.
        \item Add more annotations for nuanced activities, such as concealed weapon detection or gun transfers between individuals.
    \end{itemize}
    \item Real-Time Optimization
    \begin{itemize}
        \item Optimize the model's inference speed for deployment in real-time surveillance systems.
        \item Integrate with edge computing devices or lightweight architectures for on-site processing in security cameras.
    \end{itemize}
    \item Multi-Person and Multi-Object Interaction
    \begin{itemize}
        \item Develop capabilities for analyzing interactions between multiple persons and objects to infer group-level threats or collaborative actions.
    \end{itemize}
    \item Exploration of Alternate Object Detection Models: Transformer-Based Approaches
    \begin{itemize}
        \item Transformer-based models use self-attention to capture long-range dependencies, improving detection accuracy in complex scenes compared to CNNs.
        \item Empirical studies suggest that transformer-based architectures outperform conventional object detection methods in scenarios characterized by high object density and occlusion.
    \end{itemize}
\end{enumerate}

\vspace{10pt}
Moreover, while our results are promising, we acknowledge the ethical considerations and potential privacy concerns associated with deploying such technologies in public spaces. Future implementations will need to carefully balance safety enhancements with respect for individual privacy and civil liberties.
\end{justify}

\section{\textbf{Ethical Statement}}
\begin{justify}
In our work of detecting the firearms using the deep learning methodology, we foresee several ethical considerations and limitations. The data, sourced from publicly available databases from internet and CCTV footage, is selected with an emphasis on privacy, ensuring that personally identifiable information is not compromised. However, the use of such data from CCTV/surveillance sources may raise privacy concerns and the possibility of data misuse, which deviates from our intent of promoting public safety and crime prevention. Our methods, which involve deep learning and computer vision, inherently have the risk of biases embedded in the training data, which may affect the accuracy of the model. We are also cautious about the impact of the false negatives and false positives, in ambiguous situations which could result in unwarranted suspicion or ignoring the real threats. Despite our efforts to incorporate comprehensive evaluation metrics, we can still see that these may not fully capture the real-world complexities and ethical concerns in using such technology. Therefore, we welcome thorough scrutiny from the scientific community to address these problems, with the aim to achieve a balanced strategy maximizing public safety while keeping ethical standards and respecting privacy.

\end{justify}

\end{spacing}

\end{document}